\documentclass[letterpaper, 10 pt, conference]{ieeeconf}  

\IEEEoverridecommandlockouts                              

\usepackage{graphicx}
\usepackage{adjustbox}
\usepackage{multicol}
\usepackage{tabularx}
\usepackage{caption}
\usepackage{amssymb}
\usepackage{cellspace}
\usepackage{hyperref}
\usepackage{cite}
\usepackage[table]{xcolor}
\usepackage{multirow}
\usepackage{xspace}
\usepackage{amsmath}
\usepackage{soul}

\newcommand{\acro}{CGDF\xspace}
\newcommand{\sediff}{SE3Diff\xspace}

\title{\LARGE \bf Constrained 6-DoF Grasp Generation on Complex Shapes for Improved Dual-Arm Manipulation
}

\author{Gaurav Singh$^{1*}$, Sanket Kalwar$^{1*}$, Md Faizal Karim$^{1}$, Bipasha Sen$^2$, Nagamanikandan Govindan$^{1}$, \\ Srinath Sridhar$^{3}$ and K Madhava Krishna$^{1}$
\thanks{$^*$Equal Contributions}
\thanks{$^{1}$Robotics Research Center, IIIT-Hyderabad}%
\thanks{$^{2}$Massachusetts Institute of Technology}%
\thanks{$^{3}$Brown University}%
}

\begin{document}

\maketitle
\thispagestyle{empty}
\pagestyle{empty}

\begin{abstract}

Efficiently generating grasp poses tailored to specific regions of an object is vital for various robotic manipulation tasks, especially in a dual-arm setup. This scenario presents a significant challenge due to the complex geometries involved, requiring a deep understanding of the local geometry to generate grasps efficiently on the specified constrained regions.
Existing methods only explore settings involving table-top/small objects and require augmented datasets to train, limiting their performance on complex objects. We propose \acro: \underline{C}onstrained \underline{G}rasp \underline{D}iffusion \underline{F}ields, a diffusion-based grasp generative model that generalizes to objects with arbitrary geometries, as well as generates dense grasps on the target regions. \acro uses a part-guided diffusion approach that enables it to get high sample efficiency in constrained grasping without explicitly training on massive constraint-augmented datasets. We provide qualitative and quantitative comparisons using analytical metrics and in simulation, in both unconstrained and constrained settings to show that our method can generalize to generate stable grasps on complex objects, especially useful for dual-arm manipulation settings, while existing methods struggle to do so.
Project page: 
{\href{https://constrained-grasp-diffusion.github.io/}{https://constrained-grasp-diffusion.github.io/}}

\end{abstract}

\section{INTRODUCTION}
Grasping is a critical robot capability serving a wide array of applications spanning industrial automation, household assistance, and beyond.
Significant progress has been made in building grasp generation methods~\cite{Mousavian_2019_ICCV,sundermeyer2021contact,mahlerDexNetDeepLearning2017a, chen2023kgnv2}. These works primarily focus on generating stable and collision-free grasps \textbf{uniformly distributed} across the object or scene to support pick and place tasks. However, they have certain limitations.

\begin{figure}[t]
    \centering    
    \includegraphics[width=\linewidth]{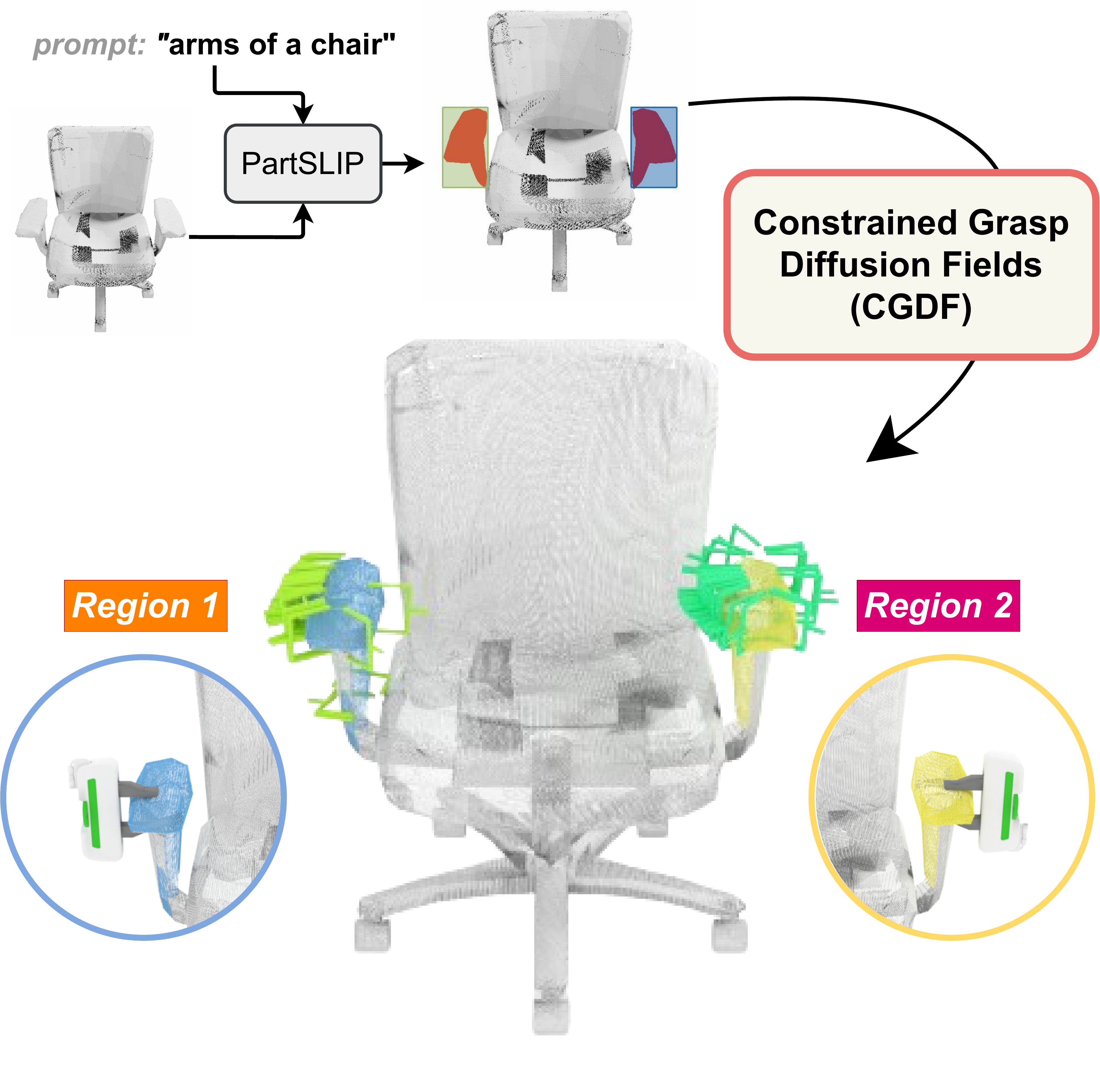}
    \caption{\small \acro: \underline{C}onstrained \underline{G}rasp \underline{D}iffusion \underline{F}ields generates dense grasps on large objects with complex shapes (like a chair), in a dual-arm setting. Given target regions (which can be generated from a text prompt using PartSLIP~\cite{liu2023partslip}), \acro uses a Part-Guided Diffusion strategy to generate sample efficient grasps on the specified regions, enabling grasping for multiple regions for improved multi-arm grasping. 
    }
    \label{fig:banner}
\vspace{-4mm}
\end{figure}

First, uniformly distributed grasps may not be practical since objects may be fragile, unwieldy, or large~\cite{da2dataset}, necessitating grasps that are concentrated more on certain parts of the object.
Second, due to suboptimal shape representations and inherent biases, such methods fail to provide dense coverage over objects with complex shapes, making them \textbf{sample-inefficient}, i.e., one needs to run the model multiple times and generate a large number of grasps to get some good grasps on the target region. 
Due to these issues, it is desirable to have a method to generate grasps \textbf{constrained} to specific regions of the object.
\textit{Given a target region, constrained grasp generation aims to generate dense, sample-efficient grasps only on the desired region}. 
Constrained grasping for complex objects also facilitates the possibility of generating grasps for more than one arm by constraining the grasps on multiple regions of interest corresponding to multiple arms. 
This thereby enables stable ``dual-arm" grasping - an area of growing interest that is yet to be addressed widely in the community.

We introduce \textbf{ \acro: \underline{C}onstrained \underline{G}rasp \underline{D}iffusion \underline{F}ields}, a novel method designed for generating constrained 6-DoF grasps tailored to complex shapes. \acro relies on an improved shape representation allowing the grasp generator to gain detailed descriptors for complex objects. Further, we propose a \textbf{part-guided diffusion} strategy to generate such grasps in a sample-efficient manner.

\begin{figure*}[t]
    \centering    \includegraphics[width=\linewidth]{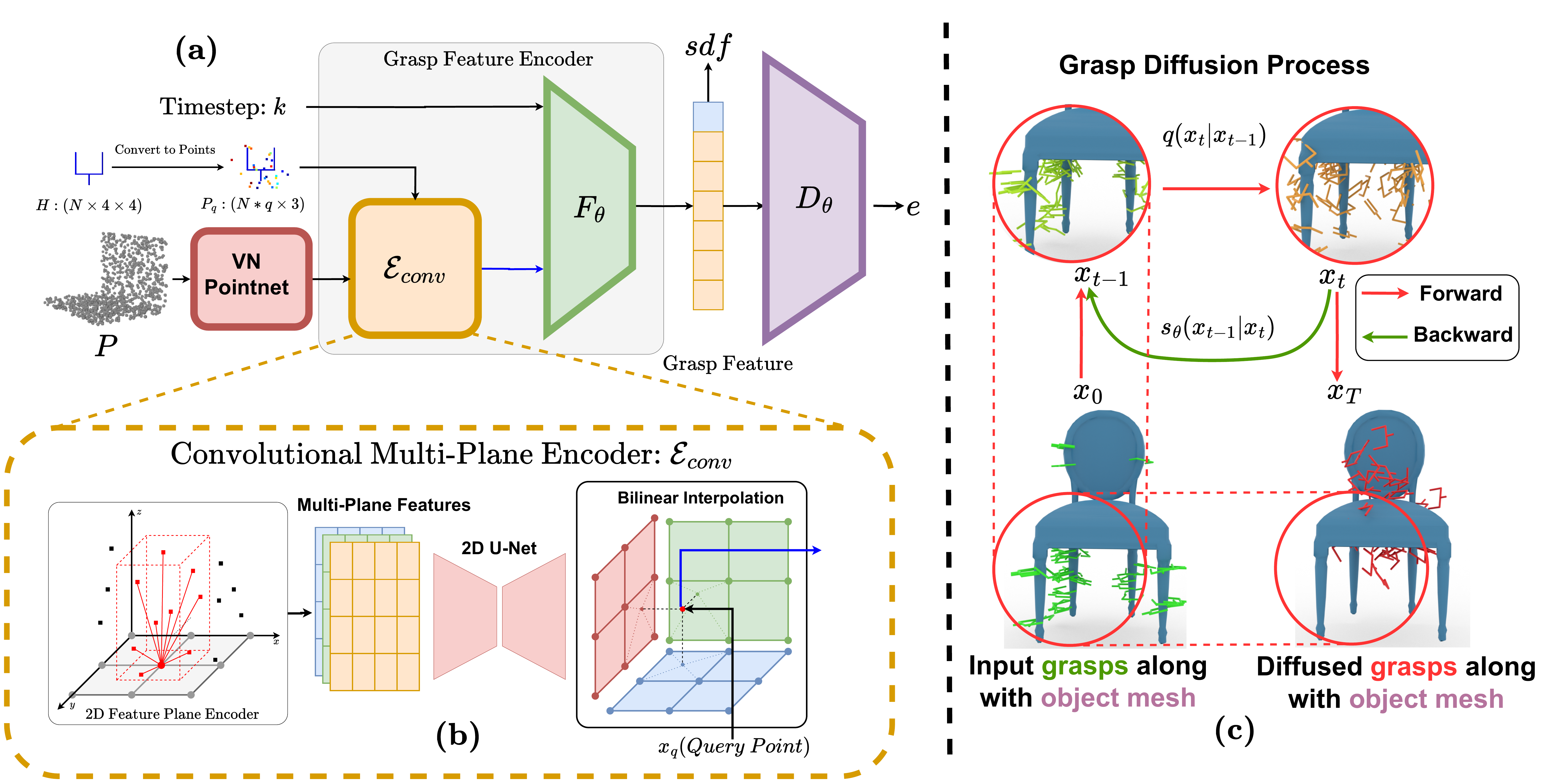} 
    \caption{\small \textbf{Overview}: 
    The figure section \textbf{(a)} shows the architecture of our proposed energy-based model $E_{\theta}$ as explained in~\ref{subsec:arch} and ~\ref{subsec:diffmodel}. During this process, the model takes as input a point cloud and the grasp pose, which is subsequently converted into a set of query points. We use a VN-Pointnet-based point cloud encoder, which generates per-point features. \textbf{(b)} shows how these features are then distilled into three 2D feature planes oriented along the XY, XZ, and YZ planes using a convolutional multi-plane encoder. For each grasp pose, feature vectors corresponding to N query points are obtained using bilinear interpolation on the feature planes. Subsequently, the grasp feature vector is derived from $F_{\theta}$ and decoded into an energy value by $D_{\theta}$. In figure section \textbf{(c)}, we show the grasp diffusion process, where grasps are diffused over the object during the forward diffusion process and denoised using the backward diffusion process.
    }
    \label{fig:main_architecture}
\end{figure*}

In this work, we assume that the regions of interest (i.e. constraints) can be easily identified using existing approaches such as today's
VLM-based affordance detection methods~\cite{liu2023partslip, langrasp, lerftogo}.
However, while these methods are good at segmenting regions based on text prompts, they lack the ability to propose grasps specific to these regions of interest thereby relying on existing uniform grasp generation methods~\cite{graspnet1billion} or simulation-based methods~\cite{1371616graspit} to sample many grasps and prune grasps based on segmented regions.
Moreover, for large objects, data-driven uniform grasping methods fail to provide dense coverage, thus becoming a bottleneck to the pipeline. Existing constrained grasping methods like~\cite{vcgs} overcome this challenge but do not generalize well to large objects with complex shapes.

We tackle the task of constrained grasp generation on arbitrary regions of large objects with complex shapes. 
Further, we showcase the efficacy of our approach in handling the complex task of dual-arm grasp generation.
The key insight of our method is that convolutional plane features \cite{convocc} that have shown significant quality improvements in fine-grained implicit 3D reconstruction \cite{deepsdf, occnet} also encode better grasp pose descriptors that improve grasp generation on complex objects. Existing general constrained grasping methods~\cite{vcgs} require expanding existing large-scale grasp datasets like ACRONYM\cite{eppner2021acronym} with constrained regions and annotated grasps on them. Due to rich local geometric features, our method, trained in an unconstrained grasping setting, can be directly applied to constrained grasping using a part-guided diffusion strategy \textit{without} losing sample efficiency, thus removing the need for such conditionally labeled datasets.

Our evaluation demonstrates the effectiveness of our method in generating stable grasps within a dual-arm setup, as evidenced by stability metrics, and in simulated environments, highlighting its practical applicability. Furthermore, conceptually our method can be extended to any number of arms, and also be helpful in multi-arm settings.
To summarize, our contributions are:
\begin{enumerate}
    \item We propose \acro, a method to generate constrained grasps on complex shapes.
    \acro is powered by convolutional plane features with the ability to store local geometries efficiently, enabling dense grasp generation on complex shapes. 
    \item We propose a novel part-conditioned generation strategy to generate sample-efficient constrained grasps without explicitly training on conditionally labeled datasets.
    \item 
    We further show the effectiveness of our approach in the complex setting of dual-arm grasping. We demonstrate that \acro outperforms existing methods in a dual-arm constrained grasp setting, showcasing the key use case of our method.
\end{enumerate}

\section{Related Work}

\subsection{Grasp Generation}
The research for single-arm grasping dealing with small objects has been well explored~\cite{Mousavian_2019_ICCV,sundermeyer2021contact,lenz2015deep,levine2018learning,mahler2017dex,morrison2020egad,zhang2019roi}. Existing single-arm grasps methods~\cite{Mousavian_2019_ICCV,sundermeyer2021contact} are trained to map the point cloud of an observed object or scene to a diverse set of grasps
and incorporate an evaluator network to handle collisions. 
The grasp generation method described in~\cite{9830843} employs a differentiable sampling procedure, guided by a multi-task optimization objective.
On the other hand,~\cite{ryu2023equivariant} presents an SE(3) equivariant energy-based model, which is able to train with a very small number of demonstrations.~\cite{chen2023kgnv2} is a 6-DoF grasp pose synthesis approach from 2D/2.5D input based on estimated keypoints.~\cite{multiselectivedual} proposed a possible dual-arm grasping strategy and a multi-stage learning method for selective dual-arm grasping using Convolutional Neural Networks (CNN) for grasp point prediction and semantic segmentation on the RGB images. However, these methods focus on generating uniform grasps for the entire object or scene, limiting their use case in constrained grasping.

\subsection{Constrained generative grasp sampling}
Constrained grasping methods have previously been explored for both single-arm~\cite{murali2021same,vcgs} and dual-arm settings~\cite{zhao2023dualafford}. However,~\cite {murali2021same,zhao2023dualafford} focus on task-oriented grasps, and are thus limited by the tasks they train on. 
We follow the setting of \cite{vcgs}, which focuses on constraining arbitrary regions. It proposes a novel constrained 6-DoF generative grasp sampler, VCGS, trained on the CONG dataset that generates dense grasps on arbitrarily specified target regions. The proposed CONG dataset is an augmentation of the ACRONYM~\cite{eppner2021acronym} dataset, which contains randomly subsampled target grasping regions along with the ground truth grasps on them. These areas can represent, for instance, semantically meaningful locations on the target object, such as the handle of a cup or the bottle cap, but can also cover the entire object. We extend the task setting of~\cite{vcgs} to a dual-arm setting, using the $DA^2$ dataset~\cite{da2dataset}. However, unlike VCGS, we do not need a conditionally labeled dataset as our model can directly generate constrained grasps without explicitly training on that objective.

\section{Background}

\underline{Convolutional Occupancy Networks~\cite{convocc}}
\label{sec:conv-occ} represent a 3D surface implicitly as a decision boundary of a neural network classifier and make a popular representation choice for fine-grained implicit 3D reconstruction \cite{deepsdf, occnet} tasks due to the translation equivariance and their ability to reconstruct finer details. It stores features in the form of 3 feature planes, each of the dimension $H\times W\times d, d$ being the feature dimension. Given a pointcloud, it projects the per-point features (from a pointcloud encoder, e.g. PointNet ~\cite{qi2017pointnet}) onto these three canonical planes (aligned with the coordinate axes) and aggregates features for each pixel in the grid using average pooling. These planes are then processed by a shared 2D U-Net to achieve translation equivariance.

\underline{Neural Descriptor Fields~\cite{ndfs}} 
\label{sec:ndf-bg}
represent point descriptors as the vector of concatenated activations of a conditional occupancy function modeled by a neural network. A query pose can then be represented by the point descriptors of a fixed query pointcloud $P_q$ transformed to that pose. This results in the descriptors of a grasp at the query pose being similar to the descriptors of the target region of that grasp.

\underline{SE(3) Diffusion Fields~\cite{se3dif}} formulate grasp pose generation as a gradient-based inverse diffusion process~\cite{gradtrain_ebm} of an SE(3) Diffusion model. This method essentially samples random grasp poses and moves them to ``low-cost" regions that represent good grasping poses. SE(3) Diffusion models offer improved coverage and representation of multimodal distributions, such as those encountered in 6DoF grasp generation scenarios. This enhancement contributes to superior and more sample-efficient performance in subsequent robot planning tasks. 

\section{\acro: \underline{C}onstrained \underline{G}rasp \underline{D}iffusion \underline{F}ields}
\label{sec:method}

Given a object point cloud $P\in\mathbb{R}^{N\times3}$ along with a target region (i.e. constraint) 
$P_t\subseteq P$, our goal is to generate $M$  parallel-jaw grasp poses $H_{i} \in SE(3), i \in[0, M)$ on $P$, such that $H_{i}$ is located on $P_t$. Here $M$ is a number that can be user-defined. Additionally, we make no assumptions regarding the granularity of the target regions, allowing $P_t$ to encompass anything from a small, localized region to the entire point cloud. 
For an n-arm setting with large objects, $n$ target regions can be selected $P_T =\{P_t^1, P_t^2 \dots P_t^n\}$. Consequently, our method must learn a robust shape representation to generalize to arbitrary shapes and complex geometries, ensuring its applicability across various constraint regions. Further, the model should generate diverse grasps with full coverage over the target region. Diffusion models have shown excellent performance in generating diverse outputs while learning the data distribution effectively~\cite{se3dif}. Thus we choose a diffusion-based architecture as our backbone.

In this section, 
we \textbf{(A)} first dive into constructing a grasp generation diffusion model.
\textbf{(B)} We then formulate a strategy to generate sample-efficient 
grasps constrained regions without explicitly training on conditionally labeled data. 
In \textbf{(C)}, we outline the architecture of our model, enabling \acro to achieve better grasps on complex shapes, with and without constrained regions.

\subsection{SE(3) Diffusion Model}
\label{subsec:diffmodel}
Our diffusion model architecture uses the formulation presented by \sediff~\cite{se3dif} for grasp generation. To adapt Euclidean diffusion models to the Lie group SE(3)~\cite{micro_lie}, \sediff works in the vector space $\mathbb{R}^6$, which is isomorphic to the \textbf{Lie algebra} $\mathfrak{se}(3)$. This allows one to apply linear algebra to an element $H$ in the Lie group SE(3). $H$ can be moved between the Lie group and the vector space using the logarithmic and exponential maps: $\text{Logmap : }SE(3)\rightarrow\mathbb{R}^6$ and $\text{Expmap : }\mathbb{R}^6 \rightarrow SE(3)$. Please refer to~\cite{se3dif} for more details. A \textbf{diffusion model in SE(3)} can then be formulated as a vector field $s_{\theta}$ that returns a vector $v \in \mathbb{R}^6$, given a query pose $H \in \text{SE(3)}$, pointcloud $P$ and the current noise scale $k$~\cite{song2019generativediff}. Formally: $v_k = s_{\theta}(H,k,P)$.

\textbf{Energy Based Model:} Following \sediff, instead of directly training a model to learn a vector field $s_{\theta}$, we construct an energy-based model $E_{\theta}$ as shown in Fig~\ref{fig:main_architecture}(a). This learns a scalar field representing the energy of the grasp distribution, as this allows the model to score the generated grasps. The energy $e_k$ of a grasp pose $H$, given a point cloud $P$ is defined as:
\begin{equation}
\label{eqn:energy}
    e_k = E_{\theta}(H,k,P)
\end{equation}
Following the notation in~\cite{micro_lie}, the derivative of a function mapping $SE(3)$ to $\mathbb{R}$ w.r.t to an SE(3) element is a vector $v\in \mathbb{R}^6$. Since $E_{\theta}: SE(3) \rightarrow \mathbb{R}$, $s_{\theta}$ can then be defined as the derivate of $e_k$ w.r.t $H$:
\begin{equation}
\label{eqn:score_gradient}
    s_{\theta}(H,k,P) = \frac{De_k}{DH}
\end{equation}
This formulation enables a \textbf{scoring mechanism} for the grasp poses based on the energy values, where lower energy corresponds to better grasps.

\textbf{Forward diffusion Process}: As shown in Fig~\ref{fig:main_architecture}(c), we first sample a perturbed data point by first moving the ground-truth grasp pose $H$ from the Lie group to the vector space using the $\text{Logmap}$ function. We then add gaussian noise $\epsilon_k \in \mathbb{R}^6$ to it, with standard deviation $\sigma_k$ for noise scale $k$, and map it back to SE(3) using the Expmap function  as follows:
\begin{equation}
\label{eqn:perturbed}
    H_k = \text{Expmap}\left[ \text{Logmap}(H) + \epsilon_k \right], \epsilon_k \sim \mathcal{N}(0,\sigma_k^2I)
\end{equation}

\textbf{Loss Function:} The diffusion model can now be trained to predict the noise given a perturbed sample by minimizing the L1 loss objective between the sampled perturbation $ \epsilon_k$ and the predicted perturbation as follows:
\begin{equation}
\label{eqn:lossfn}
    \mathcal{L}_{\text{diff}} = ||s_{\theta}(H_k,k,P) - \epsilon_k||
\end{equation}

\textbf{Inverse Diffusion Step:} Following Se3diff~\cite{se3dif}, the inverse diffusion step is formulated as an adapted version of the Euclidean Langevin MCMC~\cite{neal2011mcmc}, as shown in Fig~\ref{fig:main_architecture}(c):
\begin{equation}
    H_{k-1} = \text{Expmap}\left(- \frac{\alpha_k^2}{2}s_{\theta}(H_k,k,P) + \alpha_k\epsilon \right)H_k
\end{equation}
where $\epsilon \in \mathbb{R}^6, \epsilon \sim \mathcal{N}(0,I)$ , $\alpha_k$ refers to a non-negative step-dependent coefficient.

\subsection{\textbf{Constrained Grasp Diffusion}}
\label{subsec:constrained}
The above diffusion model trains on an unconstrained grasping objective and thus generates grasps covering the entire object pointcloud $P$. However, if the model is able to learn a rich enough shape representation to model local geometries, we can use the model trained on unconstrained grasping to generate constrained grasps without retraining, by using the energy values predicted by $E_{\theta}$ as explained below.

\begin{figure}[t]
    \centering
    \includegraphics[width=\linewidth]{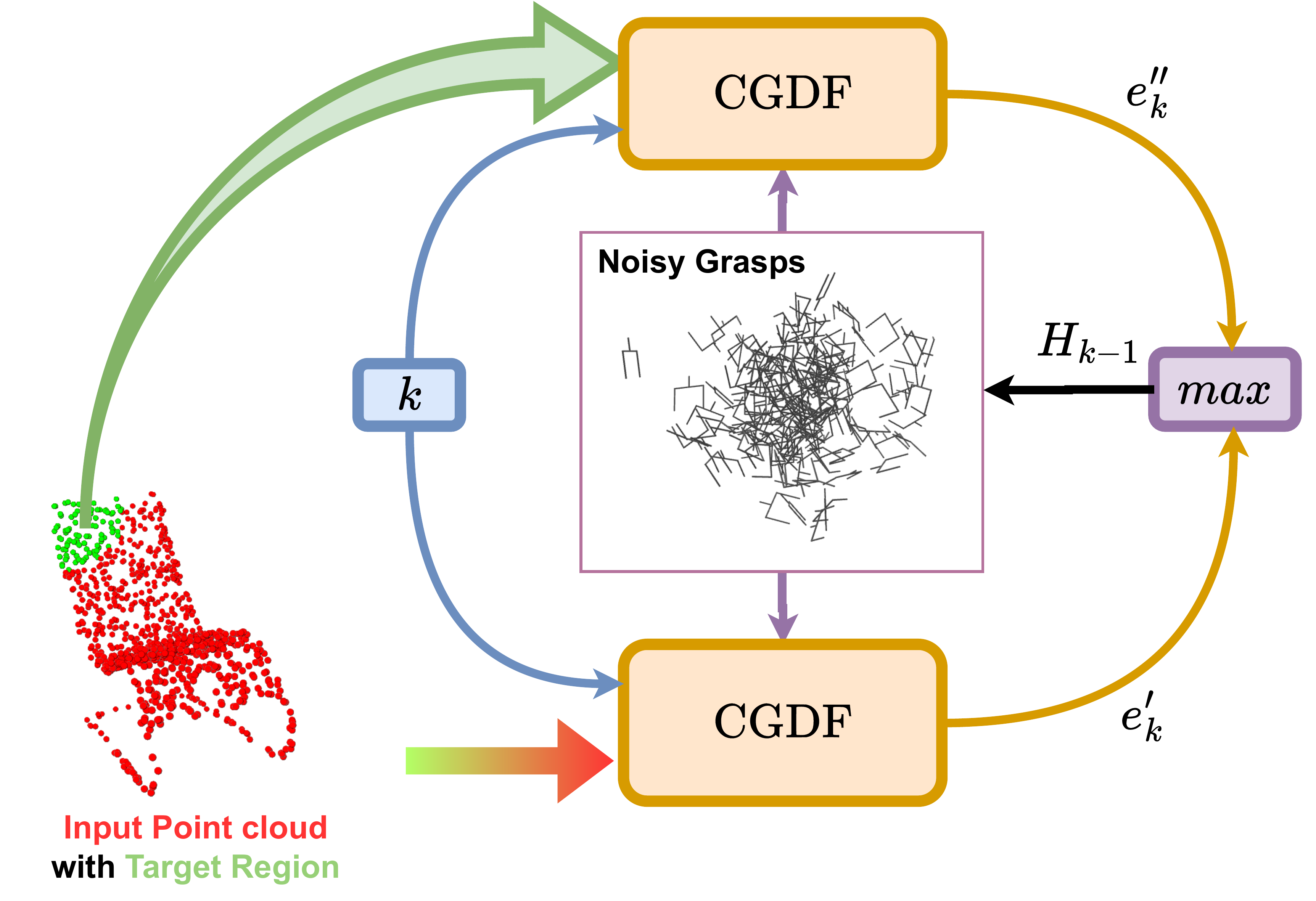}
    \caption{\small \textbf{Part-Guided Diffusion strategy:} Using 2 instances of \acro, conditioned on full pointcloud and the target region, we show how our part-guided diffusion works.}    
    \label{fig:algo}
\end{figure}

Assume we have an object pointcloud $P$ and a target region pointcloud $P_t$.
Since $E_{\theta}$ is an EBM, stable grasp poses have lower energies as compared to invalid or colliding grasps.
Therefore, if we treat $P_t$ as a stand-alone pointcloud, then a grasp pose $H_t$ on it is stable if its energy at $k=0$ ($e'' = E_{\theta}(H_t, 0, P_t)$) is low (using Eqn~\ref{eqn:energy}). Now, we can also calculate the energy of the $H_t$ giving $P$ as the context instead of $P_t$, that is, $e' = E_{\theta}(H_t, 0, P)$. Notice that $e'$ is low only if $H_t$ is also a stable grasp on $P$. Thus, $H_t$ is a \textbf{stable constrained grasp} on $P$ if:
$$
    e' < \delta \And e'' < \delta
$$
where $\delta$ is the user-defined energy threshold for valid grasps. This threshold depends on the training conditions of the model and can only be defined empirically by looking at the range of energy values returned by $E_{\theta}$.

Based on the above insight, we formulate a method to generate sample efficient constrained grasps using a \textbf{part-guided diffusion} strategy, illustrated by Fig.~\ref{fig:algo}. We modify Equation \ref{eqn:energy} as follows:
\begin{equation}
\label{eqn:max}
    e_k = \text{max}(e_k',e_k'')
\end{equation}
where,
$$
e_k' = E_{\theta}(H_k,k, P); e_k'' = E_{\theta}(H_k,k, P_t)
$$
Equation \ref{eqn:score_gradient} can now be written as:
$$
    s_{\theta}(H_k, k, \{P,P_t\}) = \frac{D\text{max}(e'_k,e''_k)}{DH_k}
$$
\begin{equation}
\label{eqn:guidance}
    \therefore s_{\theta}(H_t, k, \{P,P_t\}) = \begin{cases}
        s_{\theta}(H_k, k , P) & \text{if } e'_k \ge e''_k \\
        s_{\theta}(H_k, k , P_t) & \text{if } e'_k < e''_k \\
    \end{cases}
\end{equation}

As shown in Equations \ref{eqn:guidance} and \ref{eqn:max}, by taking the maximum value of the energies, we essentially guide the grasp from a random pose to a stable configuration near the target region. During the inverse diffusion step, if the grasp $H_k$ is near the constrained region but collides with the full object or is unstable, the energy $e'_k$ is higher, which makes $H_k$ move to a more stable pose. Conversely, if the grasp is stable but is far from the constrained region, then $e''_k$ is higher, and the grasp moves closer to the constrained region. Eventually, the grasp moves to a pose where the energies for both $P$ and $P_t$ are low, i.e. the grasp is on the constrained region as well as stable.

\subsection{\textbf{Model Architecture}}
\label{subsec:arch}
Our model architecture is inspired by~\cite{ndfs} and~\cite{se3dif}, which show that the joint learning of surface reconstruction and grasping generates rich grasp features. However, these methods only use a global point cloud embedding as the conditional vector.
This approach works for smaller objects where smaller details do not significantly affect grasp poses but fails to give good results when working with larger objects with complex shapes. 
Since part-guided diffusion is heavily dependent on the energy values, it requires that the EBM $E_{\theta}$ learns a shape representation that is expressive enough to distinguish between arbitrary shapes at a fine-grained level.
To achieve this required fidelity, we build on top of the architecture of~\cite{se3dif} and incorporate local geometric features inspired from~\cite{convocc}, as shown in Fig~\ref{fig:main_architecture}. 
Given an input pointcloud P$\in \mathbb{R}^{N \times 3}$, we first use a VN-Pointnet~\cite{deng2021vector} combined with a convolutional plane encoder to generate three canonical feature planes. Given a query point $x_q$, we obtain the point-feature vector $z_{x_q}$ at that point using by sampling from the feature planes using bilinear interpolation. The feature encoder $F_{\theta}$ is an MLP that outputs the sdf along with the feature vector at that point as follows:
$$f_{x_q}^k = \{sdf_{x_q},\psi_{x_q}^k\} = F_{\theta}(z_{x_q} \oplus x_q, k)$$ 
where $\oplus$ means concatenation.
The grasp pose $H \in SE(3)$ is represented as $P_H = H\cdot P_q$, where $P_q \in \mathbb{R}^{N\times3}$ is a \textit{fixed query pointcloud}. The pose descriptors of $H$ can then be obtained as $f_H^k = \bigoplus_{i=0}^N [f_{x_i}^k ], x_i \in P_H$. The decoder is an MLP that takes the grasp pose descriptors and outputs the energy of the grasp, $e_k = D_{\theta}(f_{H}^k)$. The model trains on the main objective as defined in Equation~\ref{eqn:lossfn}, along with an auxiliary loss to predict the SDF as mentioned in~\cite{se3dif}.

\begin{figure}[t]
    \centering
    \includegraphics[width=\linewidth]{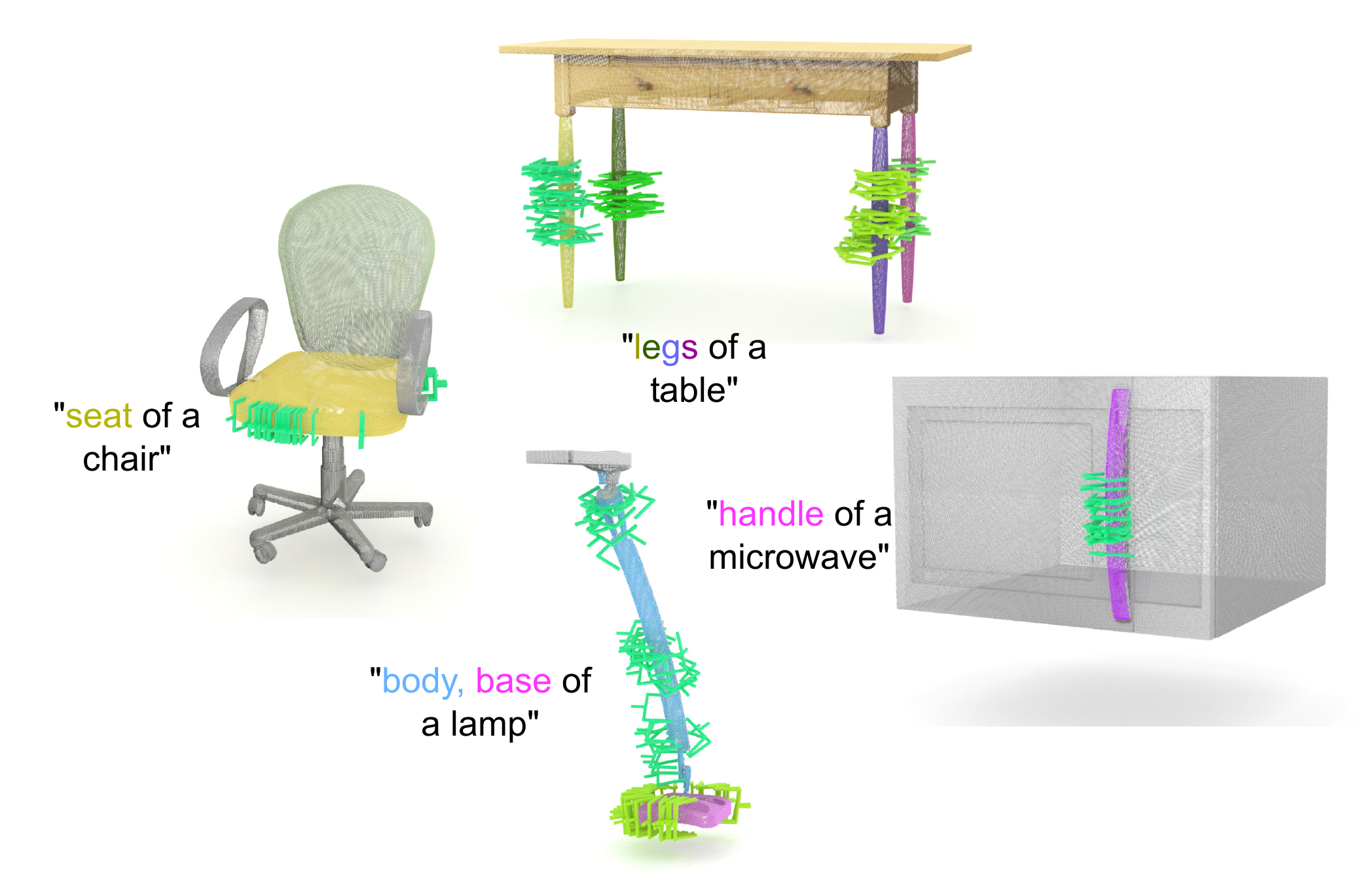}
    \caption{\small Qualitative results of PartSLIP to generate the constrained region based on given text prompts followed by \acro to generate grasps on the proposed regions.}  
    \label{fig:partslip}
\end{figure}
\section{Experiments}
\label{sec:expts}
In this section, we evaluate \acro on Constrained and Unconstrained Generation on complex and large objects objects, in a dual-arm setting. Although these grasps can be evaluated from single-arm stability, it is not practical to grasp large complex objects just using a single arm, and an evaluation from a dual-arm perspective gives us better implications on the performance of these models. 

We sample pointclouds (\#points=1000, irrespective of the size of the object) for all objects and use them for constrained and unconstrained grasping. For constrained grasping, we generate two target randomly sampled target regions by first getting two query points using farthest sampling and getting $k$ nearest neighbors, each for the 2 points. We keep $k=100$ for our experiments. Even though $k$ is constant, we notice that based on the geometry and the location of the query point, these constrained regions have coverage from a small section (e.g. on a planar region) to a large part of the object, such as in the case of thin structures like legs of a table or the neck of a guitar. We then generate constrained grasps using the method described in Section \ref{sec:method}. For unconstrained grasping, we take as input the object point cloud as a whole rather than constrained regions. For dual-arm grasp generation, we prune out grasps where the 2 grippers are close to each other and calculate dual-arm stability metrics. We also show ablation results showcasing the effects of how each design decision helps our method. 

\textbf{Datasets:} We train and evaluate all models on grasps and objects from the DA$^2$ Dataset ~\cite{da2dataset}, which consists of dual-arm grasps sampled on arbitrary object meshes taken from ~\cite{shapenet}. The meshes used in this datasets are taken from the ShapeNetSem dataset, which are scaled up relative to the grasp, making learning shape prior relatively challenging due to the large number of classes and fewer instances per class. The ACRONYM dataset ~\cite{eppner2021acronym}, also uses meshes from the same set but is catered towards smaller tabletop objects, and scales the large objects down to table-top/toy size, whereas DA$^2$ keeps large objects like chairs, lamps, etc., in their original scale. Since the baseline models require a large-scale dataset consisting of object point clouds, along with successful grasps on randomly sampled targets, we augment the DA$2$ dataset using the same procedure as described in ~\cite{vcgs} Section V. We use the original DA$^2$ dataset without augmentation to train our model. 
We further show qualitative results emphasizing the generalization capabilities of our model by combining our method with a language-based part segmentation model PartSLIP~\cite{liu2023partslip} on the PartNetE Dataset~\cite{partnet,Xiang_2020_SAPIEN}. As shown in Fig~\ref{fig:banner} and~\ref{fig:partslip}, we segment out the part using a text prompt (e.g.\textit{``handle of a microwave"}). We then use our model with the part-guided diffusion strategy (trained on DA$^2$) to generate grasps on the region.

\textbf{Baselines:} For constrained grasping, we compare with VCGS ~\cite{vcgs}, a SoTA-constrained grasp sampling method. We sample grasp pairs using VCGS-Sampler, and output the highest scoring grasps using VCGS-Evaluator.
For unconstrained grasp generation, we further compare with vanilla SE(3)-DiF model, based on which our architecture was built.

\begingroup
\begin{table}[t]
\label{tab:main_results}
    \centering
    \adjustbox{max width=\textwidth}{
    \begin{tabular}{|c|r|c|c|}
        \hline
        Metric & Methods  & Unconstrained  & Constrained \\
         \hline
         \multirow{3}{*}{FC (\%)$\uparrow$}  & VCGS & 3.28   &  3.96\\
        & SE3Diff  & 14.22 & - \\
        & \acro (\textbf{Ours}) & \cellcolor{lightgray!25}\textbf{43.51} & \cellcolor{lightgray!25}\textbf{44.8}\\
         \hline
         \multirow{3}{*}{GSR (\%)$\uparrow$} & VCGS  & 41.01 & 43.36\\
         & SE3Diff & 46.2 & - \\
         & \acro (\textbf{Ours}) & \cellcolor{lightgray!25}\textbf{60.3} & \cellcolor{lightgray!25}\textbf{60.88} \\
         \hline
         \multirow{2}{*}{TG (\%)$\uparrow$} & 
         VCGS & 100 & 76.2 \\
         & \acro (\textbf{Ours}) & 100 & \cellcolor{lightgray!25}\textbf{91.86} \\
         \hline
    \end{tabular}}
    \caption{\small Quantitative comparison \acro with SE3Diff \cite{se3dif} and VCGS \cite{vcgs}  in unconstrained and constrained setting.  Force Closure (FC) and Grasp Success Rate(GSR) are explained in Sec~\ref{sec:expts}. Target Grasp (TG)\% refers to the ratio of grasps on the target regions as explained in  Sec~\ref{sec:expts}.}
    \label{tab:main_results}
\end{table}
\endgroup

\begin{figure*}[t]
    \centering
    \includegraphics[width=0.95\linewidth]{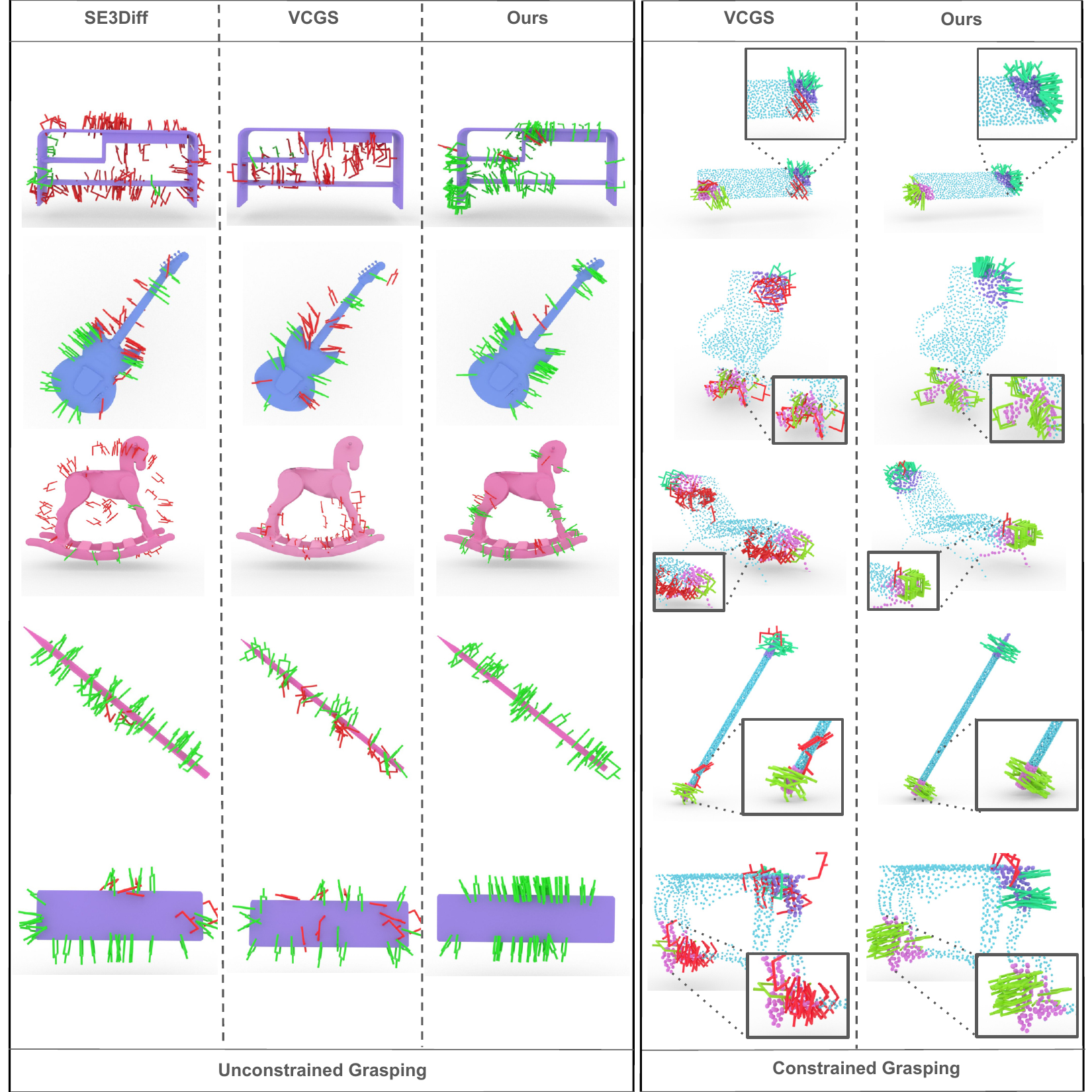}
    \caption{\small Qualitative comparison of \textit{unconstrained} and \textit{constrained} grasp generation in a dual-arm setting on \acro and the baselines: VCGS~\cite{vcgs} and SE3Diff\cite{se3dif}. All the baselines perform well for objects with simple shapes (like planar or elongated shapes). However, for relatively complex geometries like chairs, instruments, etc. \acro generates dense, constrained and unconstrained grasps, whereas the baselines struggle to do so. The green grasps are non-colliding, and the red grasps are colliding. \textbf{Zoom in for a better experience.}}
    \label{fig:qualitative}
\end{figure*}

\textbf{Metrics}:

We evaluate the performance of these models using three metrics: Force Closure, Grasp Success Rate and Target Grasps. Force Closure serves as a stringent criterion, accounting for both stability and collisions between the object and the gripper in the generated dual-arm grasps. On the other hand, the Grasp Success Rate reflects a more practical assessment, simulating scenarios typical of physical interactions within an Isaac Gym environment~\cite{isaac}. Target Grasps measures the percentage of grasps that are generated near the target region.

\textit{Target Grasps (TG)}: To test if a grasp is within the target area, we take the minimum distance between the grasp center and all points in the target area and prune grasps that have a distance greater than $6\text{cm}$, which is the assumed finger length for the DA$^2$ dataset.

\textit{Force Closure (FC)}: We evaluate the stability of the generated dual-arm grasps using the force closure metric as defined in ~\cite{da2dataset, liu2021synthesizing}, which defines a set of contact forces (4 in our case, 2 for the gripper of each hand) to be force closure if 
\begin{equation}
    GG' \succcurlyeq \epsilon I_{6 \times 6};\text{ and }
    \|Gc\| < \delta'
\end{equation}
where $G$ is the grasp matrix,, $c$ is the axis of the friction cone, $S$ is the object surface and $ x_i \in S$. Please refer to ~\cite{liu2021synthesizing} for more details.
To get the contact points, we first create a gripper mesh, transform it to the predicted pose $H$, and prune out any grasps that intersect with the object mesh. This is a relatively hard constraint as grasps that collide even slightly are pruned out. We then sample rays inwards from the gripper fingers and find the intersection with the ground truth mesh of the object, along with contact normals from the mesh faces. We then define the FC metric as the percentage of grasps that satisfy force closure out of the total number of predicted grasps (including the ones that collide).

\textit{Grasp Success Rate (GSR)}: We evaluate the grasp success rate in a simulation environment. Similar to the setting in ~\cite{vcgs}, we use two free-floating \textit{Franka Emika Panda} grippers with extended fingers (6cm), to grasp a free-floating object so as to not bias the evaluation with the grasp reachability or approach directions. We place the two grippers at the two grasp poses and simultaneously close the grippers until either the fingers touch or the object is grasped. We then turn on gravity and lift the two grippers up a considerable distance. After this, if the distance between the grippers and the object is less than the span of the object, it means that the object has not fallen down, and we consider that configuration as a success. We run all simulation experiments on the Isaac Gym Simulator ~\cite{isaac}.

\textbf{Results}
As shown in Table~\ref{tab:main_results}, \acro outperforms the existing methods on all metrics as well as sample efficiency. The existing methods rely only on the global shape representation. This makes them prone to errors where the object includes finer details like multiple thin structures. As seen from the qualitative outputs shown in Fig ~\ref{fig:qualitative}, VCGS-Sampler is able to generate grasps close to the constrained region.
However, it fails to model the fine geometry of the region and generates multiple grasps that slightly touch the objects causing collisions and lowering the FC metric drastically. VCGS-Evaluator, which is also trained with a global shape representation, fails to model the accurate shapes well and thus classifies many of the slightly colliding candidate grasps as positive. However, these slight collisions do not have a drastic effect on the grasp success as such grasps cause the object to shift slightly, while still being successful. Even though SE3Diff generates relatively fewer colliding grasps, it still achieves almost the same level of grasp success as VCGS, showcasing the limitations of using global embeddings.

\textbf{Ablations:}
\acro has 2 key design decisions: (1) Convolutional Plane features and (2) Part-Guided Diffusion. We evaluate the contribution of both of these decisions, which enable our network to improve grasp performance.

\textit{(A) Convolutional Plane features}:
Convolutional plane features~\cite{convocc} represent local geometries efficiently by distilling features from existing pointcloud encoders like VN-PointNet, as shown in Figure~\ref{fig:main_architecture}. Since we learn a joint representation for shapes and grasps (by converting them into points )\cite{ndfs}, the features of a grasp pose $H$ are similar to the target region of the grasp. Thus, we analyze how well this 3D representation encodes the object geometry. To do this, we compare vanilla SE3Diff\cite{se3dif} with our model. Since both these models generate SDF as an auxiliary task, we analyze the role of convolutional features using Chamfer's Distance (CD) \footnote{\textcolor{blue}{\href{https://pdal.io/en/stable/apps/chamfer.html}{https://pdal.io/en/stable/apps/chamfer.html}}} between pointclouds sampled from the predicted SDF and the ground truth pointcloud. As shown in Table~\ref{tab:ablations},\acro significantly improves the reconstruction quality compared to SE3Diff, showcasing the effectiveness of Convolutional plane features.

\textit{(B) Part Conditioned Diffusion}:
It is possible to generate grasps on constrained regions with \acro without using the part-guided strategy, by employing a naive albeit sample-inefficient approach mentioned in Section~\ref{subsec:constrained}, which is to simply generate grasps on the target region pointcloud $P_t$ and threshold grasps based on energy values. In this experiment, we try to study the effect of part-guided diffusion over this naive method in terms of sample efficiency. For this, we measure the percentage of grasps left after the thresholding step for grasps generated with and without part-guided diffusion (w/o PD). Since the bounds of energy are not fixed \cite{gradtrain_ebm}, we set the energy threshold $\delta$ such that the FC metric of the grasps from both the models is in a similar range as in Table~\ref{tab:main_results}. As shown in Table~\ref{tab:ablations}, part-guided diffusion generates \textbf{2.5 times} times more valid grasps than the baseline (w/o PD).

\begingroup
\begin{table}[t]
\label{tab:ablations}
    \centering
    \adjustbox{max width=\textwidth}{
    \begin{tabular}{|c|r|c|c|}
        \hline
        Modification & Metric & Ours  & Modified \\
        \hline
        w/o Conv & CD $\downarrow$  &  \textbf{14.04}  &  60.21 \\
        \hline
        w/o PD & SE (\%)$\uparrow$  &  \textbf{93.6}  & 37.4 \\
         \hline
    \end{tabular}}
    \caption{\small Ablation study: We show the contribution of different components of our model. In w/o Conv, we evaluate the quality shape representations of SE3Diff \cite{se3dif} and \acro by comparing the reconstructions of both the models. Chamfer's Distance (CD) is scaled by $10^4$. SE (sample efficiency) refers to the percentage of grasps remaining after the energy thresholding step.
    }
    \label{tab:ablations}
\end{table}
\endgroup

\section{CONCLUSIONS}
Existing data-driven grasp generation methods primarily focus on uniformly generating stable and collision-free grasps across target objects, limiting their effectiveness in scenarios requiring constrained grasps on complex geometries. To address this limitation, we introduce \acro (Constrained Grasp Diffusion Fields), a diffusion-based grasp generative model capable of generalizing to objects with arbitrary geometries and generating dense grasps on specified regions. 
\acro utilizes a part-guided diffusion approach, enabling high sample efficiency in constrained grasping without the need for extensive constraint-augmented datasets. Through qualitative and quantitative evaluations in both unconstrained and constrained settings, we demonstrate \acro's ability to generate stable grasps on complex objects for dual-arm manipulation, surpassing existing methods.

\textbf{Future work} could involve possible extensions to scenarios involving more than two arms, further improving the representation,  potentially expanding \acro's application to a broader range of robotic manipulation tasks.






\bibliographystyle{IEEEtran}
\bibliography{eg}

\end{document}